\documentclass[letterpaper, 10 pt, conference]{ieeeconf}  

\IEEEoverridecommandlockouts                              

\usepackage{graphicx} 
\usepackage{amsmath,amsfonts}
\usepackage{algorithm}
\usepackage{algpseudocode} 
\usepackage{array}
\usepackage[caption=false,font=normalsize,labelfont=sf,textfont=sf]{subfig}
\usepackage{textcomp}
\usepackage{stfloats}
\usepackage{url}
\usepackage{verbatim}
\usepackage{graphicx}
\usepackage{cite}
\usepackage{xcolor}
\usepackage{booktabs}
\usepackage{tabularx}
\usepackage{subfig}
\title{\LARGE \bf RGB: RL Guided Whole-Body MPPI for Humanoid Control}
 \author{Yunsoo Seo, Sol Choi, Euncheol Im, Myo Taeg Lim, Yisoo Lee%
 \thanks{*This work was supported by the National Research Foundation of Korea (NRF) grant funded by the Korea government(MSIT) (RS-2024-00339632, RS-2025-25448259).}
 \thanks{Y. Seo, S. Choi, and Y. Lee are with the Center for Humanoid Research, Korea Institute of Science and Technology (KIST), 02792 Seoul, South Korea. Y. Seo is now with University of Texas at Austin, Austin, Texas 78712, USA. S. Choi is also with the Department of Mechanical Engineering, Yonsei University, 03722, Seoul, South Korea}
 \thanks{M. T. Lim is with the School of Electrical Engineering, Korea University, 02841 Seoul, South Korea}
 \thanks{\textit{Corresponding author: Yisoo Lee} {\tt\small yisoo.lee@kist.re.kr}}
}
\date{December 2025}

\begin{document}

\maketitle
\thispagestyle{empty}
\pagestyle{empty}

\begin{abstract}
Humanoid robots require whole-body controllers that are both robust and precise in contact-rich environments. While deep reinforcement learning (RL) achieves robust stability, its behavior is tightly coupled to the training objective and command interface, making it difficult to add new feedback objectives without retraining. 
In this study, we propose an RL guided whole-body model predictive path integral (MPPI) framework that acts as an add-on feedback controller on top of a pretrained RL policy. Instead of using RL policy as the final controller, we use it as a sampling prior that biases MPPI rollouts toward dynamically feasible behaviors.
Task objectives are specified through modular MPPI cost terms, and MPPI closes the loop by continuously correcting the RL prior online to satisfy these objectives without retraining the policy. 
Simulations on a 29-DoF Unitree G1 humanoid in MuJoCo demonstrate stable high-rate control (average 280~Hz).
The proposed method improves task-level precision over a pure RL baseline under the same command interface.
This is achieved by correcting systematic drift during straight walking and tracking additional whole-body reference signals imposed through the cost.
\end{abstract}

\section{Introduction}


 Humanoid robots are expected to perform sophisticated, complex, and human-like motions while maintaining dynamic balance and ensuring safety in uncertain, contact-rich environments. To achieve these capabilities, prior research has primarily followed two major paradigms: model-based control~\cite{zhang2025mujocoilqr} and learning-based approaches centered on reinforcement learning (RL)~\cite{10075792}. Model-based methods provide interpretability and stability guarantees grounded in system dynamics, whereas learning-based approaches offer adaptability and the ability to handle high-dimensional behaviors. Nevertheless, both paradigms exhibit inherent limitations, making it difficult to develop a unified whole-body control framework that simultaneously achieves high precision, robustness, and strong generalization across diverse tasks and environments.

 In particular, deep RL has demonstrated impressive whole-body locomotion skills with strong stability and adaptability~\cite{haarnoja2024soccer, radosavovic2024realworld, li2024realizing}. By enabling humanoid robots to experience diverse and challenging scenarios in simulation, RL facilitates the acquisition of robust behaviors. However, enhancing robustness through extensive domain randomization and disturbance training often introduces a trade-off, where motion precision and fine-grained control performance may be compromised.

Furthermore, while RL has achieved remarkable performance in specific tasks, deploying it as a general-purpose whole-body controller remains challenging. Learned policies are inherently shaped by their training objectives and command interfaces.
For instance, incorporating additional objectives such as explicit swing foot clearnace modulation or pelvis height regulation to a locomotion policy trained for planar velocity tracking typically accept limited command inputs (e.g., base linear velocity and yaw rate)~\cite{margolis2024rapid}, requires task specific reward redesign and retraining. This results in significant development time and limited scalability across tasks.

Recent efforts have leveraged human motion data to learn generalized whole-body motion tracking policies via RL~\cite{ze2025twist2, luo2025sonic}. While these approaches broaden the behavioral repertoire, achieving both rapid, highly dynamic motion generation and fine-grained precision in control remains an unresolved challenge. Moreover, RL-based policies do not naturally enforce strict constraint satisfaction, such as torque limits, which is critical for reliable and safe humanoid operation.

 Model-based methods, particularly recent whole-body model predictive control (MPC), offer a structured approach by optimizing composite task objectives online while explicitly enforcing system constraints~\cite{bishop2025linearwalking, molnar2025whole}. Sampling-based MPC methods, such as Model Predictive Path Integral (MPPI)~\cite{williams2016icra}, are especially appealing for contact-rich humanoid locomotion, as they can optimize non-smooth objectives through parallel rollouts without requiring analytic derivatives~\cite{williams2017model, alvarezpadilla2024wbmppi, TRO_ours, humanoid_mine}. More recently, physics engine-based predictive control frameworks, including MuJoCo MPC (MJPC)~\cite{howell2022predictive, alvarez2025real}, have further improved the practicality of whole-body predictive control by leveraging high-fidelity forward dynamics and efficient rollout computation~\cite{xue2025full}.

Despite these advantages, applying sampling-based MPC to high-DoF humanoids remains challenging. In particular, vanilla MPPI is highly sensitive to initialization and sampling distributions. Achieving feasible rollouts and stable performance often requires substantial manual tuning and carefully designed proposal distributions, and performance can still degrade under rapidly changing contact conditions.

These complementary strengths and limitations motivate a hybrid approach that preserves the robustness of learned locomotion while enabling explicit objective shaping and feedback refinement through predictive control. In this study, we propose an RL guided whole-body predictive control framework that uses a pretrained RL policy as a sampling prior within a physics engine-based MPPI. Crucially, our method can be viewed as an add-on, plug-and-play controller, which the base RL policy remains unchanged and continues to provide robust locomotion, while MPPI adds generalized whole-body feedback objectives through modular cost terms.

The main contributions of this work are as follows:
\begin{enumerate}
    \item We propose an RL guided whole-body MPPI framework that acts as an add-on feedback controller for a pretrained RL policy.
    \item We establish a modular objective formulation that enables flexible task-level command augmentation without modifying or retraining the learned policy.
    \item We implement an asynchronous, physics-engine-based MPPI on a 29-DoF humanoid and achieve real-time operation with an average effective update rate of $\sim$280~Hz in simulation.
    \item We demonstrate improved task-level precision over a pure RL baseline under identical command inputs, while preserving robust locomotion from the RL prior.

    
    

\end{enumerate}

\section{Background}
\subsection{Model predictive path integral (MPPI)}
MPPI~\cite{williams2017model,williams2018information} is a sampling-based stochastic optimal control method that optimizes a control sequence by evaluating a population of perturbed rollouts over a finite prediction horizon.
At each control step, the method samples control perturbations, simulates the corresponding trajectories, and aggregates them through an importance-weighted update. 
In this work, MPPI serves as the model-based predictive controller, which is integrated with a learned policy as a sampling prior.

We briefly review the standard formulation of MPPI. Consider a discrete-time dynamical system
\begin{align}
    x_{t+1} = f(x_t, u_t + \delta u_t), \quad \delta u_t \sim \mathcal{N}(0, \nu I),
    \label{mppi_dynamics}
\end{align}
where $x_t$ denotes the system state at time $t$, $f(\cdot)$ is the discrete-time system dynamics.
The control input is decomposed into a nominal signal $u_t$ and an gaussian exploration noise term $\delta u \sim \mathcal{N}(0,\nu I)$ that drives the sampling process with covariance $\nu I$. The stochastic exploration around $u$ is what later allows us to bias the sampling distribution using RL.

The control objective is expressed as the minimization of a finite-horizon cost functional
\begin{align}
    S(\mathbf{x}) 
    = \phi(x_T) 
    + \sum_{t=0}^{T-1} l(x_t, u_t),
    \label{mppi_cost}
\end{align}
where $\mathbf{x} = [x_t,\dots,x_T]$ is the state trajectory, $\phi(x_T)$ is a terminal cost, and $l(x_t, u_t)$ is a running cost that encodes the task objective and soft constraints.

In the path-integral derivation of MPPI, a likelihood-ratio term appears when relating the controlled dynamics to an uncontrolled reference process. This change of measure naturally introduces a quadratic penalty on the control and leads to an augmented instantaneous cost of the form
\begin{align}
    \tilde{l}(x,u,\delta u)
    &= l(x,u)
    + \frac{(1 - \nu^{-1})}{2}\,\delta u^\mathsf{T} R \,\delta u \nonumber \\
    &\quad + u^\mathsf{T} R \,\delta u 
    + \tfrac{1}{2} u^\mathsf{T} R \,u,
    \label{running_cost_mppi}
\end{align}
where $R$ is a positive-definite weighting matrix. The additional terms arise from the injected noise $\delta u$ and recover the usual quadratic effort penalty $\tfrac{1}{2}u^\mathsf{T}R u$ known from classical optimal control.

Given this modified running cost, each sampled trajectory $n \in \{1,\dots,N\}$ accumulates a cost-to-go $\tilde{S}_n(\mathbf{x})$. MPPI then updates the control sequence by computing an importance-weighted average of the sampled perturbations:
\begin{align}
    u_t \leftarrow u_t +
    \frac{
        \sum_{n=1}^{N}
        \exp\!\Bigl(
            -\tfrac{1}{\lambda}
            \bigl(\tilde{S}_n(\mathbf{x}) - \tilde{S}_{\min}\bigr)
        \Bigr)\,
        \delta u_{t,n}
    }{
        \sum_{n=1}^{N}
        \exp\!\Bigl(
            -\tfrac{1}{\lambda}
            \bigl(\tilde{S}_n(\mathbf{x}) - \tilde{S}_{\min}\bigr)
        \Bigr)
    },
    \label{mppi_discrete_update}
\end{align}
where $\delta u_{t,n}$ is the perturbation applied at time index $t$ for the $n$-th rollout, $\tilde{S}_{\min}$ is the minimum cost among all sampled trajectories, and $\lambda$ is the inverse temperature parameter controlling how sharply the weighting focuses on low-cost trajectories. Intuitively, rollouts with smaller cumulative cost receive exponentially larger weights, so the updated control is biased toward perturbations that produced more desirable outcomes.

This sampling-based update has several properties that are particularly relevant for high-dimensional humanoid control.
First, the update relies solely on rollout evaluations and does not require derivatives of the dynamics or cost, allowing the incorporation of complex nonlinear models or physics engines. 
Second, rollouts are independent and can be evaluated in parallel, enabling efficient computation for real-time predictive control.
These properties motivate the hybrid framework introduced in the next section, which leverages derivative-free rollouts and parallel computation for whole-body control.


\subsection{Limitations of Vanilla MPPI for Humanoids}
Despite its simplicity and derivative-free nature, vanilla MPPI can be brittle for high-dimensional humanoid locomotion. 
First, MPPI performance depends strongly on the sampling distribution, including the choice of covariance and the mean, which is often set by warm-starting from the previously optimized sequence. 
Poorly chosen sampling statistics can lead to inefficient exploration, large variance in rollout costs, and convergence to suboptimal local solutions~\cite{underactuated, mohamed2025toward}. 
Second, in contact-rich tasks, dynamically feasible trajectories may occupy only a narrow subset of the state–action space.
As a result, many sampled rollouts may violate implicit feasibility (e.g., unstable contact sequences), wasting samples and degrading real-time performance. 
This sensitivity is characteristic of nonconvex trajectory optimization in high-dimensional systems~\cite{kelly2017trajopt}.
Finally, although physics engine-based rollouts enable derivative-free optimization, a large number of samples may still be required to reliably improve the control sequence under tight real-time constraints, particularly when discontinuous contact transitions are present.

\section{Proposed Method}

To address the sensitivity of vanilla MPPI to sampling statistics and initialization in high-dimensional humanoid control, we propose a RL guided whole-body MPPI framework that integrates a learned motion prior with physics engine-based predictive optimization.

\subsection{Overview of the RL Guided MPPI framework}

\begin{figure*}[t]
    \centering
    \includegraphics[width=\textwidth]{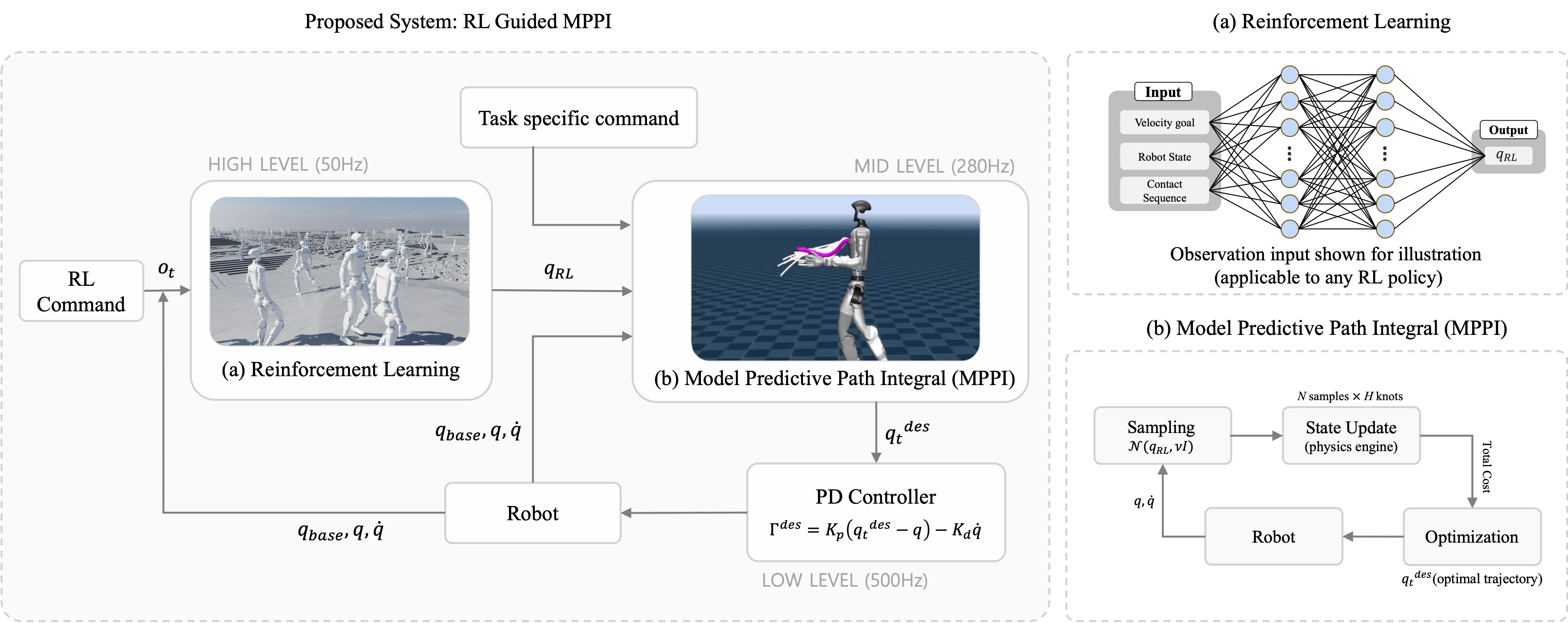}
    \caption{Overview of the proposed RL guided whole-body MPPI framework. A pretrained RL policy provides a nominal motion prior for sampling, and knot based MPPI refines it using physics rollouts and modular cost terms as an add-on feedback controller. The refined command is applied immediately upon completion of planning in an asynchronous receding-horizon loop, and a low-level PD controller tracks the joint targets.}
    \label{RL_WholebodyMPPI_framework}
\end{figure*}
Fig.~\ref{RL_WholebodyMPPI_framework} shows overall structure of the proposed hybrid control framework. The framework combines (a) a pretrained RL policy that provides a motion prior and (b) a knot-parameterized MPPI that refines the prior using physics-consistent rollouts.

The overall control stack operates at three different rates. 
First, the RL policy runs at 50\,Hz and outputs a nominal joint position reference $q_{\mathrm{RL}}$. 
Second, the MPPI refines this prior in an asynchronous receding-horizon loop. Once planning finishes, it outputs an updated desired joint command using $q_{\mathrm{RL}}$ as the mean of the sampling distribution, yielding an average effective update rate of $\sim$280\,Hz in simulation with CPU-based parallel rollouts.
Finally, a low-level joint PD controller runs at 500\,Hz, tracking the MPPI-refined desired states ($q_t^{des}$) and computing the corresponding desired joint torques $\Gamma^{des}$.
These update rates are implementation choices rather than fixed design constraints, and both the RL policy and MPPI modules can operate at higher frequencies depending on computational resources.
Further details can be found in the following subsections and Algorithm~\ref{alg:rl_mppi_simple}.




\subsection{RL Policy-Based Sample Trajectory Generation}
\subsubsection{RL policy}
An RL policy provides a nominal whole-body joint position that serves as a structured sampling prior for MPPI. 
We employ a pretrained humanoid policy, a command velocity tracking locomotion policy, as the prior for the proposed RL guided MPPI. 
The framework is agnostic to the specific RL algorithm, training procedure, or reward design. 
Any learned policy that outputs a nominal whole-body command with a compatible action interface can be used. 
Importantly, the RL policy is trained as a standalone controller, without considering the presence of MPPI, and is assumed to be sufficiently generalized for direct deployment in real robot control. 
Note that the proposed method does not require training or modification of the policy to integrate it into the predictive control framework.

\subsubsection{RL policy as a sampling prior}
\label{sec:rl_prior}
At each control step, the controller observes the current RL observation $o_t$ and the RL policy $\pi_{\mathrm{RL}}(\cdot)$ outputs a nominal joint position reference $q_{RL}$. This reference is utilized as the mean of the sample generator in MPPI, around which Gaussian perturbations are applied.

Note that the command inputs used by the RL policy are shared with MPPI.
The same command vector contained in the RL observation $o_t$ is used to generate the nominal prior $\bar{\mathbf z}=\pi_{\mathrm{RL}}(o_t)$ and is also used consistently during MPPI rollouts so that refinement is performed around the intended commanded behavior of the prior. 
Additional task objectives are introduced only through MPPI cost terms as task-space references, enabling closed-loop refinement without modifying the RL command interface.


\subsubsection{Knot-based interpolation}
\label{Knot-based interpolation}
Directly optimizing a full control sequence over the horizon leads to a high dimensional search space.
Thus, we employ knot-based sampling approach~\cite{tao2026sampling} which parameterize the MPPI control sequence using $H$ knots.
Knot-based interpolation enables the evaluation of continuous and extended trajectories using only a small number of sampled control sequences, thereby improving computational efficiency and exploration effectiveness.

Specifically, we treat the knot vector $\mathbf z=[z_1^\top,\dots,z_H^\top]^\top \in \mathbb{R}^{H\times J}$ as the optimization variable, where $J$ is the number of actuated joint DoFs and each $z_t\in\mathbb{R}^J$ corresponds to a joint-position target at a knot time. 
At each MPPI update, we sample $H$ knot sequences around the RL mean. Then, we can compose the following equation.
\begin{equation}
\mathbf z^{(n)}=\pi_{\mathrm{RL}}(o_t)+\epsilon^{(n)}, \qquad \epsilon^{(n)}\sim\mathcal{N}(0,\nu I),
\label{eq:rl_knot_prior_sampling}
\end{equation}
where policy input $o_t$ denotes the observation used during RL training,
$\epsilon$ represents the sampled joint position target increments at knot nodes, and $\nu$ controls the exploration variance around the policy prior. 
The above equation reconstructs a high-resolution control sequence from the knot representation using cubic interpolation, 
$U^{(k)} = \mathrm{InterpolateKnots}(\mathbf z^{(k)})$.
This reduces the effective sampling dimension from $T\times N \times J$ to $H \times N \times J$ and yields temporally smooth commands, since the full rate time sequence control input is reconstructed by interpolating between knots.
The resulting control sequence is then evaluated through rollouts to compute the corresponding trajectory and cost.

\subsection{Physics Engine-based Optimal Solution}
\subsubsection{Physics Engine-based Rollout}
MPPI injects Gaussian perturbations according to~\eqref{eq:rl_knot_prior_sampling} around the nominal sequence and samples $N$ candidate trajectories in a low-dimensional knot space with $H$ knots. 
Each candidate is evaluated through forward simulation in the physics engine, and an importance-weighted update is computed based on the resulting task costs.
Generated perturbed control propagate rollouts through the forward dynamics
\begin{align}
    x_{h+1}^{(n)} = f\!\left(x_{h}^{(n)}, u_{h}^{(n)}\right),
    \qquad h=0,\dots,H-1.
    \label{eq:rl_prior_rollout}
\end{align}

In our framework, the physics engine acts as a rollout oracle. Given the current state and a candidate control sequence, it performs forward simulation—including contact dynamics—and returns the resulting trajectory and cost. 
This process does not require analytic Jacobians, explicit dynamics derivations, or backward sensitivity propagation. 
The derivative-free nature of this approach is particularly advantageous for contact-rich humanoid locomotion, where intermittent contacts and non-smooth transitions make local linearization and gradient-based methods unreliable.

MPPI naturally supports parallel computation, as each rollout is independent. 
Accordingly, physics engine-based predictive control frameworks including MJPC evaluate large batches of candidates trajectories in parallel.
By combining learning-guided sampling with physics engine-supported parallel rollouts, the proposed framework enables fast and scalable predictive optimization for contact-rich humanoid control.

\subsubsection{Cost-based Optimal Solution Computation}
\label{sec:Cost-based Optimal Solution Computation}
For each rollout $n$, we compute the finite-horizon trajectory cost $S^{(n)}$ with equations \eqref{mppi_cost} and \eqref{eq:rl_prior_rollout}.
\begin{align}
    S^{(n)} = \phi\!\left(x_{H}^{(n)}\right) + \sum_{h=0}^{H-1} \ell\!\left(x_{h}^{(n)}, u_{h}^{(n)}\right),
    \label{eq:proposed_traj_cost}
\end{align}
The terminal and running costs $\phi$ and $\ell$ together constitute a multi-objective cost function, allowing task objectives and constraints to be composed and weighted according to the control requirements.
Then the normalized importance weights are described as follows:
\begin{equation}
w^{(n)}=
\frac{
        \sum_{n=1}^{N}
        \exp\!\Bigl(
            -\tfrac{1}{\lambda}
            \bigl(\tilde{S}_n(\mathbf{x}) - \tilde{S}_{\min}\bigr)
        \Bigr)\,
        \epsilon^{(n)}
    }{
        \sum_{n=1}^{N}
        \exp\!\Bigl(
            -\tfrac{1}{\lambda}
            \bigl(\tilde{S}_n(\mathbf{x}) - \tilde{S}_{\min}\bigr)
        \Bigr)
    },
\end{equation}
Using the importance weights $w^{(n)}$, we compute the refined knot vector $\mathbf z^\star=\bar{\mathbf z}+\sum_{n=1}^{N} w^{(n)}\epsilon^{(n)}$.
We then interpolate $\mathbf z^\star$ to a dense desired joint-position sequence and set $q_t^{des}$ to its first element.
Only $q_t^{des}$ is executed, and the optimization is repeated at the next time step during the receding-horizon.

\subsubsection{Control Input}
\label{sec:Control Input}
Finally, the desired joint torques are computed using a joint-space PD controller and applied to the robot actuators:
\begin{equation}
\Gamma^{des} = K_{p}(q^{des}_{t}-q) - K_{d}\dot{q},
\end{equation}
where $q$ and $\dot{q}$ are the measured joint state vectors.
The PD gain matrices $K_{p}$ and $K_{d}$ are constant diagonal matrices and are identical to those used during policy training.



```latex
\begin{algorithm}[t]
\caption{RL-Guided Knot-Based MPPI Overview}
\label{alg:rl_mppi_simple}

\begin{algorithmic}[1]

\State Initialize nominal knots
\Statex \hspace{\algorithmicindent}
$\bar{\mathbf{z}} \gets \mathbf{0}$

\While{the system is running}

    \State Read state $x_t$

    \If{the RL policy is updated at 50\,Hz}
        \State
        $\bar{\mathbf{z}} \gets \pi_{\mathrm{RL}}(o_t)$
        \Comment{Sec.~\ref{sec:rl_prior}}
    \EndIf

    \If{the planner computation is complete}

        \For{$j = 1$ to $J$}

            \For{$n = 1$ to $N$}

                \State Sample knot noise
                \Statex \hspace{\algorithmicindent}
                $\epsilon^{(n)}
                \sim \mathcal{N}(0,\nu I)$

                \State
                $\mathbf{z}^{(n)}
                \gets
                \bar{\mathbf{z}}+\epsilon^{(n)}$
                \Comment{Sec.~\ref{sec:knot_interpolation}}

                \State
                $U^{(n)}
                \gets
                \operatorname{InterpolateKnots}
                (\mathbf{z}^{(n)})$

                \State
                $(X^{(n)},S^{(n)})
                \gets
                \operatorname{RolloutAndCost}
                (x_t,U^{(n)})$

            \EndFor

            \State
            $\{w^{(n)}\}_{n=1}^{N}
            \gets
            \operatorname{ImportanceWeights}
            \left(
            \{S^{(n)}\}_{n=1}^{N},
            \lambda
            \right)$
            \Comment{Eq.~\ref{eq:proposed_traj_cost}}

            \State
            $\bar{\mathbf{z}}
            \gets
            \bar{\mathbf{z}}
            +
            \sum_{n=1}^{N}
            w^{(n)}\epsilon^{(n)}$

        \EndFor

        \State
        $q_t^{\mathrm{des}}
        \gets
        \operatorname{FirstControl}
        \left(
        \operatorname{InterpolateKnots}
        (\bar{\mathbf{z}})
        \right)$

    \EndIf

    \If{the PD controller is updated at 500\,Hz}
        \State Apply PD control to track $q_t^{\mathrm{des}}$
        \Comment{Sec.~\ref{sec:control_input}}
    \EndIf

\EndWhile

\end{algorithmic}
\end{algorithm}
```

\subsection{Design Rationale}
\subsubsection{Decoupling Learning from Task Specification}
The RL policy is trained with a reward function that may not be fully aligned with the objectives required at deployment time. 
Our framework explicitly decouples policy learning from task specification. 
The RL policy is used solely to shape the sampling distribution and does not act as the control law itself. 
Task objectives and constraints are instead imposed through modular MPPI cost terms. 

As a result, new tasks can be introduced by modifying the MPPI cost function without retraining or fine-tuning the RL policy. 
This separation enables flexible task composition while preserving the structured motion prior provided by the learned policy.

\subsubsection{RL Guided Sampling and Sample Efficiency}
Centering the MPPI sampling distribution on the learned policy $\pi_{RL}$ biases rollouts toward dynamically feasible whole-body behaviors. 
This significantly improves sample efficiency in high-dimensional humanoid control, where naive exploration can result in a large number of infeasible or unstable trajectories.

By leveraging a policy that has already learned stabilizing locomotion or motion patterns, MPPI exploration is concentrated in a structured region of the state–action space, reducing sensitivity to initialization and improving convergence.

\subsubsection{Implicit Contact Handling via Physics Rollouts}
Contact-rich feasibility is enforced implicitly through physics engine rollouts during MPPI evaluation. 
In our implementation, contact handling and constraint-consistent state propagation are provided by the simulator during forward simulation. 
We do not introduce additional hand-crafted contact constraint penalties in the cost function. 

Instead, infeasible contact behaviors naturally incur higher rollout costs due to unstable state evolution, allowing MPPI to reject them through importance weighting.

\section{Simulation results}
\subsection{System Setup}
We evaluate the proposed RL guided whole body MPPI in MuJoCo using a 29 DoF Unitree G1 humanoid model.
Rollout evaluation was performed using MJPC, leveraging CPU-based parallel computation.
All validations were conducted on a workstation equipped with an Intel Core i9-14900KF CPU, 32 GB RAM, and an NVIDIA GeForce RTX 4070 Ti GPU.

For the evaluation of the proposed method, a humanoid walking policy that can track command velocity is trained and utilized. While our framework is agnostic to the specific RL algorithm and policy architecture, we employ a policy trained independently in IsaacLab using proximal policy optimization (PPO)\cite{schulman2017proximal}. The policy was trained as a standalone controller for direct deployment, without considering integration with the proposed MPPI framework.
The reward consisted of command-tracking terms (linear and yaw velocity), regularization terms encouraging smooth actions and joint motions, stability/uprightness and joint-limit penalties, and gait-related terms (e.g., foot air-time and symmetry gait phase).
The policy outputs desired joint positions for the 12 lower-body DoFs. Observation includes base angular velocity, projected gravity vector, velocity command, joint positions, joint velocities, previous action, and gait phase.

For the knot-based MPPI, we use $N=128$ rollouts with a horizon of $0.02$ s, a control timestep of $dt=0.002$ s (500 Hz), $H=2$ spline knot points, and Gaussian exploration noise with standard deviation $\sigma=0.2$.
In all experiments, task objectives are specified by augmenting the MPPI stage cost in equation~\eqref{eq:proposed_traj_cost} with task-specific terms, while keeping the pretrained RL policy unchanged.

\subsection{Results}

Training a robust whole-body RL policy typically requires substantial reward tuning, and the resulting command interface is often limited. Consequently, incorporating additional objectives outside the original interface usually requires reward redesign and retraining. 

To address this limitation, we evaluate task augmentation with the proposed method.
The pretrained locomotion policy is kept unchanged and used as a robust motion prior, while MPPI augments the behavior by adding task objectives through cost terms and refining the prior in closed loop. This closed-loop refinement compensates systematic tracking errors online, improving task-level precision while preserving the robust locomotion behavior of the learned policy.

As representative tasks, we consider straight-walk drift suppression and base-height regulation for a squat motion. The straight-walk task regulates lateral position and heading to mitigate drift induced by symmetry-breaking bias under a zero yaw command. The squat task evaluates whether an explicit whole-body posture objective can be imposed through cost augmentation alone.

\paragraph{Straight-walk drift suppression}
We first evaluate walking velocity tracking under the same command interface used for RL training, commanded $x$ and $y$ velocity and  angular velocity around $z$-axis. In practice, learning a symmetry consistent locomotion policy is often challenging even when trained with symmetric observations and rewards, the resulting policy can exhibit a persistent left-right bias due to stochastic optimization, contact discontinuities.
As a consequence, a pure RL baseline may fail to straight walk under a zero angular velocity command, drifting or gradually turning despite, the yaw reference is zero (${\psi}_{\mathrm{des}}=0$).
To isolate the benefit of the proposed controller, we use a pretrained RL policy that exhibits this straight-line tracking deficiency and verify whether MPPI refinement can suppress the bias without retraining. 
In this experiment, the same locomotion command $(v_x^{des}, v_y^{des}, \psi^{des})=(1,0,0)$ is provided to both the RL policy and the MPPI rollouts, so the nominal behavior remains that of the original velocity-tracking policy. 
MPPI then augments this behavior by adding an extra task objective through the cost term $\ell$, forming a task-space feedback loop that reduces drift while preserving robust locomotion.

Specifically, we add a straight-walk objective that penalizes both lateral drift and heading drift:
\begin{equation}
\ell =
\beta_y \, \| y_{\text{base}} - y^{\text{des}} \|^2
+ \beta_{\psi}\, \| \psi_{\text{base}} - \psi^{\text{des}} \|^2,
\end{equation}
Where $y_{\text{des}}$ is set to the initial base lateral position and $\beta_y, \beta_{\psi}$ is a scalar weight used as 25.
Here, we only use the running cost $\ell$ accumulated over the finite-horizon, and we do not include an explicit terminal cost, $\phi\!\left(x_{H}^{(n)}\right)$. No additional cost terms are used beyond those defined in $\ell$. When $\ell$ is not evaluated for planning, the system reverts to the pretrained RL policy alone.

\begin{figure}[t]
    \centering
    \includegraphics[width=0.48\textwidth]
    {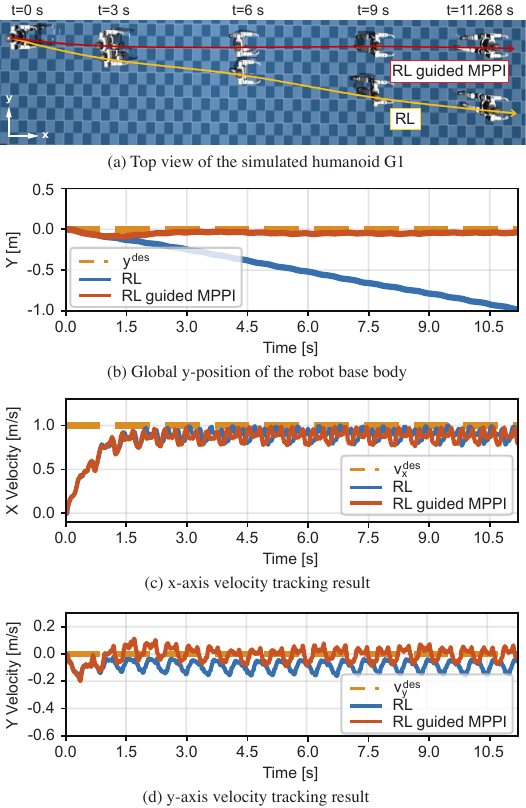}
    \caption{Straight walking performance comparison between pure RL and RL guided MPPI.}
    \label{figure_straight_walk}
\end{figure}

Fig.~\ref{figure_straight_walk} summarizes straight-walking performance over a 11.265~s trial. 
The proposed RL guided MPPI significantly suppresses drift accumulation, reducing the RMSE of the base lateral position error from 0.339~m for the pure RL policy to 0.022~m.
This indicates that the added straight-walk objective introduces a task-space closed-loop refinement that compensates systematic left--right bias in the pretrained policy, enabling near-straight locomotion while retaining the robust walking behavior provided by the RL prior. 
Fig.~\ref{figure_straight_walk}(c) further shows that the forward velocity tracking accuracy is comparable between the two methods, with RMSE values of 0.773 for RL guided MPPI and 0.806 for pure RL. 
In contrast, the lateral velocity tracking in Fig.~\ref{figure_straight_walk}(d) improves noticeably, where RL guided MPPI achieves an RMSE of 0.021~m/s compared to 0.046~m/s for pure RL. 
These results support that incorporating task-space feedback through MPPI improves motion precision, particularly in drift-related components, without degrading the nominal velocity-tracking performance of the learned policy.


\paragraph{Squat task}
To demonstrate task augmentation beyond the RL command interface, we add a base-height objective for a squat motion through cost composition. Specifically, we augment the MPPI stage cost with a base-height tracking term,
\begin{equation}
\ell = \beta_{z}\,\| z_{\text{base}} - z^{\text{des}}(t)\|^2,
\label{eq:cost_squat}
\end{equation}
where $z_{\text{base}}$ is the robot base height, $z^{\text{des}}$ is a piecewise-linear height reference, and $\beta_{z}=100$.

\begin{figure}[t]
    \centering
    \includegraphics[width=0.48\textwidth]{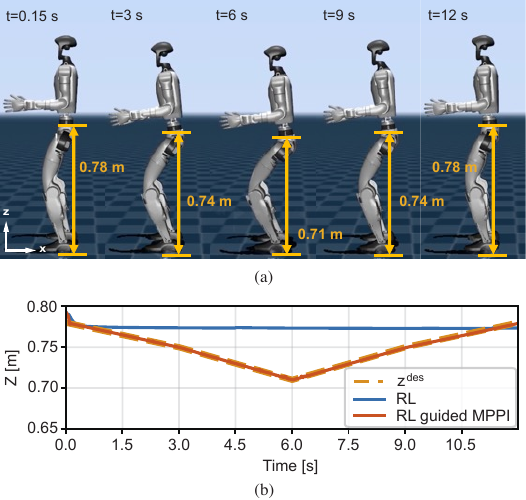}
    \caption{Squat task performance via base height cost augmentation. (a) Left view snapshots of the simulated Humanoid G1. (b) Base-height tracking with a piecewise-linear reference $z^{\mathrm{des}}$ (0.71--0.78~m). The RL baseline remains near its nominal height since $z$ is not a part of the policy command interface.
    }
    \label{squat_height}
\end{figure}


Fig.~\ref{squat_height}(b) shows that RL guided MPPI tracks the prescribed base-height reference over the entire motion, including the transition regions between descent and ascent. 

Unlike the straight-walk scenario where the command is expressed in planar task coordinates $(v_x^{des}, v_y^{des}, \psi^{des})$, the squat task introduces an additional vertical objective that the pretrained RL policy cannot directly command or regulate. 
Consistent with this, the RL only curve in Fig.~\ref{squat_height}(b) remains near its nominal base height instead of following $z^{\mathrm{des}}(t)$. 
In contrast, our proposed method closes a task-space feedback loop in the $z$ direction and achieves accurate base-height regulation without retraining.

Representative snapshots are shown in Fig.~\ref{squat_height}(a). Starting from the nominal upright posture produced by the pretrained locomotion policy, RL guided MPPI adjusts the lower-body joint targets to drive the base height toward the time-varying reference while maintaining balance. The snapshots illustrate the descent phase, where the controller increases knee and hip flexion to lower the pelvis, and the ascent phase, where it reverses this coordination to recover the original height. Importantly, these posture changes emerge from cost-based refinement around the RL prior rather than from any explicit squat command in the RL interface, demonstrating that the proposed add-on controller can synthesize a whole-body posture adaptation to satisfy a vertical task objective.








\section{Conclusion}
We presented an RL guided whole-body MPPI framework for humanoid control, where a pretrained RL policy is used as a sampling prior and task objectives are specified through a modular MPPI cost evaluated by physics engine rollouts. 
By refining the policy prior online using knot-parameterized MPPI, the proposed add-on controller improves task-level precision in real-time while maintaining the robust locomotion provided by the pretrained policy.
Furthermore, despite relying on CPU-based parallelism, our asynchronous implementation achieves an average effective update rate of 280~Hz in simulation, which GPU-based rollouts could further increase the update rate.
In simulation on a 29-DoF Unitree G1 model in MuJoCo, the method corrected a common RL failure mode in straight walking under the same command interface by reducing lateral drift, and it enabled task augmentation without retraining by incorporating additional objectives such as base-height controlling shaping via cost composition.

This work highlights the benefits of decoupling learning from task specification. Viewed as an add-on feedback controller, our MPPI module can be attached to a pretrained RL policy to retain its robust locomotion prior while injecting precise, whole-body objective shaping through modular cost design. This plug-and-play structure allows the same learned controller to be reused across tasks: new objectives can be introduced by composing or reweighting cost terms, without modifying the RL policy or its command interface.

While the proposed framework provides a strong plug-and-play way to add precise whole-body objectives on top of robust RL locomotion, its behavior is naturally anchored to the pretrained policy prior used for sampling. Extending the framework to support larger behavioral changes will be an exciting direction, for example by incorporating multiple priors (skill libraries), adaptive proposal distributions, or switching/blending mechanisms. Future work will also focus on real-hardware deployment, improving robustness to modeling and contact-estimation errors, and establishing reliable synchronization and state-estimation pipelines.




\bibliographystyle{IEEEtran} 
\bibliography{refs}          

@INPROCEEDINGS{10075792,
   author={Chen, Yizhu},
   booktitle={2023 IEEE 3rd International Conference on Power, Electronics and Computer Applications (ICPECA)}, 
   title={Walking problem of bipedal humanoid robot: comparison between model-based and learning-based method}, 
   year={2023},
   volume={},
   number={},
   pages={1629-1633},
   doi={10.1109/ICPECA56706.2023.10075792}}

@inproceedings{bishop2025linearwalking,
  title={The Surprising Effectiveness of Linear Models for Whole-Body Model-Predictive Control},
  author={Bishop, Arun L and Alvarez-Padilla, Juan and Schoedel, Sam and Sow, Ibrahima Sory and Chandrachud, Juee and Sharma, Sheitej and Kraus, Will and Park, Beomyeong and Griffin, Robert J and Dolan, John M and others},
  booktitle={2025 IEEE-RAS 24th International Conference on Humanoid Robots (Humanoids)},
  pages={1--7},
  year={2025},
  organization={IEEE}
}

@article{molnar2025whole,
  title={Whole-Body Inverse Dynamics MPC for Legged Loco-Manipulation},
  author={Molnar, Lukas and Cheng, Jin and Fadini, Gabriele and Kang, Dongho and Zargarbashi, Fatemeh and Coros, Stelian},
  journal={IEEE Robotics and Automation Letters},
  year={2025},
  publisher={IEEE}
}

@article{howell2022predictive,
  title={Predictive sampling: Real-time behaviour synthesis with mujoco},
  author={Howell, Taylor and Gileadi, Nimrod and Tunyasuvunakool, Saran and Zakka, Kevin and Erez, Tom and Tassa, Yuval},
  journal={arXiv preprint arXiv:2212.00541},
  year={2022}
}

@article{zhang2025mujocoilqr,
  title={Whole-Body Model-Predictive Control of Legged Robots with MuJoCo},
  author={Zhang, John Z and Howell, Taylor A and Yi, Zeji and Pan, Chaoyi and Shi, Guanya and Qu, Guannan and Erez, Tom and Tassa, Yuval and Manchester, Zachary},
  journal={arXiv preprint arXiv:2503.04613},
  year={2025}
}

@inproceedings{alvarez2025real,
  title={Real-time whole-body control of legged robots with model-predictive path integral control},
  author={Alvarez-Padilla, Juan and Zhang, John Z and Kwok, Sofia and Dolan, John M and Manchester, Zachary},
  booktitle={2025 IEEE International Conference on Robotics and Automation (ICRA)},
  pages={14721--14727},
  year={2025},
  organization={IEEE}
}

@article{williams2017model,
  title={Model predictive path integral control: From theory to parallel computation},
  author={Williams, Grady and Aldrich, Andrew and Theodorou, Evangelos A},
  journal={Journal of Guidance, Control, and Dynamics},
  volume={40},
  number={2},
  pages={344--357},
  year={2017},
  publisher={American Institute of Aeronautics and Astronautics}
}

@article{williams2018information,
  title={Information-theoretic model predictive control: Theory and applications to autonomous driving},
  author={Williams, Grady and Drews, Paul and Goldfain, Brian and Rehg, James M and Theodorou, Evangelos A},
  journal={IEEE Transactions on Robotics},
  volume={34},
  number={6},
  pages={1603--1622},
  year={2018},
  publisher={IEEE}
}

@inproceedings{williams2016icra,
  title     = {Aggressive driving with model predictive path integral control},
  author    = {Williams, Grady and Drews, Paul and Goldfain, Brian and Rehg, James M. and Theodorou, Evangelos A.},
  booktitle = {2016 IEEE International Conference on Robotics and Automation (ICRA)},
  year      = {2016},
  pages     = {1433--1440},
  doi       = {10.1109/ICRA.2016.7487277}
}

@article{li2024realizing,
  title   = {Realizing full-body control of humanoid robots},
  author  = {Li, Guangliang and Gomez, Randy},
  journal = {Nature Machine Intelligence},
  volume  = {6},
  pages   = {990--991},
  year    = {2024},
  doi     = {10.1038/s42256-024-00891-x}
}

@article{radosavovic2024realworld,
  title   = {Real-world humanoid locomotion with reinforcement learning},
  author  = {Radosavovic, Ilija and Xiao, Tete and Zhang, Bike and Darrell, Trevor and Malik, Jitendra and Sreenath, Koushil},
  journal = {Science Robotics},
  volume  = {9},
  pages   = {eadi9579},
  year    = {2024},
  doi     = {10.1126/scirobotics.adi9579}
}

@article{haarnoja2024soccer,
  title   = {Learning agile soccer skills for a bipedal robot with deep reinforcement learning},
  author  = {Haarnoja, Tuomas and others},
  journal = {Science Robotics},
  volume  = {9},
  pages   = {eadi8022},
  year    = {2024},
  doi     = {10.1126/scirobotics.adi8022}
}

@misc{alvarezpadilla2024wbmppi,
  title         = {Real-Time Whole-Body Control of Legged Robots with Model-Predictive Path Integral Control},
  author        = {Alvarez-Padilla, Juan and Zhang, John Z. and Kwok, Sofia and Dolan, John M. and Manchester, Zachary},
  year          = {2024},
  eprint        = {2409.10469},
  archivePrefix = {arXiv},
  primaryClass  = {cs.RO},
  url           = {https://arxiv.org/abs/2409.10469}
}

@article{ze2025twist2,
    title   = {TWIST2: Scalable, Portable, and Holistic Humanoid Data Collection System},
    author  = {Yanjie Ze and others},
    journal = {arXiv preprint arXiv:2511.02832},
    year    = {2025}
}

@article{luo2025sonic,
    title   = {SONIC: Supersizing Motion Tracking for Natural Humanoid Whole-Body Control},
    author  = {Zhengyi Luo and others},
    journal = {arXiv preprint arXiv:2511.07820},
    year    = {2025}
}

@ARTICLE{TRO_ours,
  author={Kim, Dongwhan and Im, Euncheol and Kim, Yujin and Lim, Myotaeg and Lee, Yisoo},
  journal={IEEE Transactions on Robotics}, 
  title={Single-Instance Sampling for Computationally Efficient and Accurate Real-Time Task Space MPPI Control}, 
  year={2025},
  volume={41},
  number={},
  pages={6327-6344},
  doi={10.1109/TRO.2025.3626660}}

@INPROCEEDINGS{humanoid_mine,
  author={Seo, Yunsoo and Kim, Dongwhan and Bak, Jaewan and Oh, Yonghwan and Lee, Yisoo},
  booktitle={2023 IEEE-RAS 22nd International Conference on Humanoid Robots (Humanoids)}, 
  title={Extremely Fast Computation of CoM Trajectory Generation for Walking Leveraging MPPI Algorithm}, 
  year={2023},
  volume={},
  number={},
  pages={1-7},
  doi={10.1109/Humanoids57100.2023.10375162}}

@inproceedings{xue2025full,
  title={Full-order sampling-based mpc for torque-level locomotion control via diffusion-style annealing},
  author={Xue, Haoru and Pan, Chaoyi and Yi, Zeji and Qu, Guannan and Shi, Guanya},
  booktitle={2025 IEEE International Conference on Robotics and Automation (ICRA)},
  pages={4974--4981},
  year={2025},
  organization={IEEE}
}

@article{tao2026sampling,
  title={Sampling Strategy Design for Model Predictive Path Integral Control on Legged Robot Locomotion},
  author={Tao, Chuyuan and Wang, Fanxin and Jiang, Haolong and He, Jia and Chen, Yiyang and Bu, Qinglei},
  journal={arXiv preprint arXiv:2601.01409},
  year={2026}
}

@article{margolis2024rapid,
  title={Rapid locomotion via reinforcement learning},
  author={Margolis, Gabriel B and Yang, Ge and Paigwar, Kartik and Chen, Tao and Agrawal, Pulkit},
  journal={The International Journal of Robotics Research},
  volume={43},
  number={4},
  pages={572--587},
  year={2024},
  publisher={SAGE Publications Sage UK: London, England}
}

@book{underactuated,
  title        = "Underactuated Robotics",
  subtitle     = "Algorithms for Walking, Running, Swimming, Flying, and Manipulation",
  howpublished = "Course Notes for MIT 6.832",
  author       = "Tedrake, Russ",
  year         = 2023,
  url          = "https://underactuated.csail.mit.edu",
}

@article{mohamed2025toward,
  title={Toward efficient MPPI trajectory generation with unscented guidance: U-MPPI control strategy},
  author={Mohamed, Ihab S and Xu, Junhong and Sukhatme, Gaurav S and Liu, Lantao},
  journal={IEEE Transactions on Robotics},
  volume={41},
  pages={1172--1192},
  year={2025},
  publisher={IEEE}
}

@article{kelly2017trajopt,
  title   = {An Introduction to Trajectory Optimization: How to Do Your Own Direct Collocation},
  author  = {Kelly, Matthew},
  journal = {SIAM Review},
  volume  = {59},
  number  = {4},
  pages   = {849--904},
  year    = {2017},
  doi     = {10.1137/16M1062569}
}

@article{schulman2017proximal,
  title={Proximal policy optimization algorithms},
  author={Schulman, John and Wolski, Filip and Dhariwal, Prafulla and Radford, Alec and Klimov, Oleg},
  journal={arXiv preprint arXiv:1707.06347},
  year={2017}
}
\end{document}